\documentclass{article}

\usepackage[final]{neurips_2019}

\usepackage[utf8]{inputenc} 
\usepackage[T1]{fontenc}    
\usepackage{hyperref}       
\usepackage{url}            
\usepackage{booktabs}       
\usepackage{amsfonts}       
\usepackage{nicefrac}       
\usepackage{microtype}      

\usepackage{amssymb}
\usepackage{amsmath}
\usepackage{mathtools}
\usepackage{bbm}
\usepackage[colorinlistoftodos,prependcaption,textsize=tiny]{todonotes}
\usepackage{subcaption}
\usepackage{bm}
\usepackage{tabularx}

\def\*#1{\mathbf{#1}}

\DeclareMathOperator*{\argmin}{\arg\!\min}

\title{Pairwise Feedback for Data Programming}

\author{%
  Benedikt Boecking\\
  Auton Lab, Robotics Institute\\
  Carnegie Mellon University\\
  Pittsburgh, PA 15213 \\
  \texttt{boecking@cmu.edu} \\
   \And
  Artur Dubrawski\\
  Auton Lab, Robotics Institute\\
  Carnegie Mellon University\\
  Pittsburgh, PA 15213 \\
  \texttt{awd@cs.cmu.edu} \\
}

\begin{document}

\maketitle

\begin{abstract}
The scalability of the labeling process and the attainable quality of labels have become limiting  factors for many applications of machine learning. The programmatic creation of labeled datasets via the synthesis of noisy heuristics provides a promising avenue to address this problem. We propose to improve modeling of latent class variables in the programmatic creation of labeled datasets by incorporating pairwise feedback into the process. We discuss the ease with which such pairwise feedback can be obtained or generated in many application domains. Our experiments show that even a small number of sources of pairwise feedback can substantially improve the quality of the posterior estimate of the latent class variable. 
\end{abstract}

\section{Introduction}
The  acquisition  of  labeled  data  to  train machine learning models is often an expensive and time consuming process. The scalability of the labeling process, as well as the attainable quality of  collected annotations, have become  the  limiting  factors  for  many  practical applications  of  machine  learning. Consequently, various research directions aim to improve the ways we obtain labels to train models. One traditional approach is active learning, a sequential learning process which performs queries to a teacher (sometimes referred to as an oracle) to gather feedback on data instances. Such queries are often membership queries in the case of classification. Of late, the machine learning community has developed new ideas and approaches to reduce required labeling efforts even further and to scale the labeling process, e.g. by incorporating richer forms of feedback such as information regarding feature importance~\citep{druck2008learning,raghavan2006active,settles2011closing,poulis2017learning,dasgupta2018learning}. 

Another avenue that has been proposed is to synthesize labels from imperfect sources by using generative models. A promising direction is to create labeled datasets programmatically, via user generated functions that annotate samples and provide noisy labels. This process is known as data programming~\citep{ratner2016data}, and it has been shown to produce state-of-the-art performance on several benchmark challenges as well as to yield an intuitive feedback mechanism to users.

We propose the use of weak pairwise feedback--such as same or different class membership--to improve the estimation of true class labels in the programmatic creation of labeled datasets. The appeal of incorporating pairwise feedback into data programming is that it ties evidence of labeling functions together across samples. As such, even samples which do not receive many or any weak heuristic labels can benefit from information that has been acquired for its noisy associated pairs. 
While pairwise feedback is often abundant, it is not used frequently in machine learning as it is a weaker form of supervision than direct class labels. However, pairwise feedback has been shown to be useful source of information in areas such as semi-supervised learning, metric learning, and active learning.

Pairwise information to enhance learning is easy to obtain in many applications. In some, meta-data can inform pairwise labels. For example, spatial or temporal proximity of samples can be used to induce pairwise linkage constraints, as in e.g.\ the analysis of spectral information from planetary observations~\citep{wagstaff2002intelligent}, or in video segmentation and speaker identification~\citep{bar2003learning}. Another example is to create pairwise labels from knowledge about functional links between proteins for protein function prediction tasks~\citep{eisenberg2000protein}. 

A different approach to creating pairwise feedback  is to  use locally accurate similarity functions, which are easy to obtain by experts in many domains. These functions provide ways to find small numbers of similar samples of the same class with good accuracy. In text classification for example, using cosine similarity on term frequency–inverse document frequency (tf-idf) vectors is known to routinely yield a good approach for finding nearest neighbors of the same class. Such metrics  can be used to gather high quality pairwise information with little noise, e.g. by constructing Mutual \textit{k}-Nearest Neighbor (MKNN) graphs using the user-supplied functions. MKNN approaches have been successfully used in various domains, including in single-cell RNA sequencing~\citep{haghverdi2018batch}.

In this paper, we show how to use noisy pairwise structures in combination with labeling functions in the data programming framework to improve the performance of latent class structure modeling. 


\section{Related Work}
\subsection{Data programming}
A recent trend in machine learning is to use generative models to synthesize training data from weak supervision sources. The idea is to model training set creation as a process, such that the true label is a latent variable which generates the observed, noisy labels. 
\cite{ratner2016data} introduce data programming, where a set of labeling functions programmatically label data. These label functions are created by annotators, usually with domain expertise. The acquired labels are noisy, but generative models can be used to learn parameters of the labeling process such as accuracies of labeling functions as well as their correlation structure. Consequently, these generative models can synthesize labels from such weak supervision sources at scale. 

In data programming, users can induce dependencies between labeling functions, such as that one `reinforces' another. In this context, \cite{bach2017learning} propose a structure estimation method to identify the generative model’s dependency structure. \cite{varma2019learning} introduce a robust PCA-based algorithm for dependency structure estimation.

Further extensions to data programming have been studied to address latent subsets in data as well as to handle multi-task problems and multi-resolution settings. \cite{varma2016socratic} explore automatic identification of latent subsets in the training data in which the supervision sources perform differently than on average. The discovery of the subsets is used to augment the structure of the generative model in data programming.  In an extension of data programming to the multi-task setting, \citet{ratner2018snorkel,ratner2018training}  assume that each labeling function returns a vector containing feedback on some subset of all possible tasks. For sequential data such as videos, \cite{sala2019multi} propose a framework for modeling weak multi-resolution sources. The approach handles weak supervision at different scales such as per-frame or per-scene labels on videos, as well as sequential correlation amongst weak label sources.

\subsection{Learning with Pairwise Information}
Supervision via pairwise information has been used to improve clustering performance in semi-supervised clustering. The pairwise information is usually available in the form of constraints--one set of \textit{must-link} pairs and one set of \textit{cannot-link} pairs--and the problem setting is generally referred to as constrained clustering or semi-supervised clustering~\citep{wagstaff2001constrained,Basu2002seed,klein2002instance}. A number of semi-supervised clustering algorithms have been developed which simultaneously adapt the underlying notion of similarity or distance~\citep{basu2004probabilistic,bilenko2004integrating,yan2006adaptive}. 

Learning with pairwise supervision has also been explored in active learning settings, often with pairwise comparisons/rankings. 
For example, \cite{parikh2011relative} study how to learn via relative feedback. The authors learn ranking functions for each attribute and subsequently build a generative model over the joint space of ranking outputs. 
\cite{xu2017noise} consider an active dual supervision classification problem with algorithms that query oracles for noisy labels and pairwise comparisons. For the latter, the feedback is provided in terms of a pairwise ranking of the likelihood of being positive instead of an absolute label assignment. The algorithm can leverage both types of oracles, direct label assignment and pairwise comparisons. This active learning scheme can be useful in application areas where pairwise comparisons are easier to obtain.  The comparison oracle is used to rank data points in order to create sets within which to obtain absolute labels. \cite{xu2018nonparametric} study active learning in a regression scenario with ordinal (or comparison) information. The authors provide theoretical guarantees and introduce an algorithm for this scenario. 

\section{Problem Setting}
In data programming~\citep{ratner2016data}, \textit{labeling functions} are heuristics which provide noisy labels for samples. A generative model is used to model the latent class variable that generates the noisy observations. The model estimates the accuracies of the labeling functions, as well as their dependencies in some cases. The distribution over the latent class variable can be used to train a noise-aware discriminative model.

In this paper, we assume a multi-class setting with $c$ classes. We are given a dataset with unknown labels $\bm Y\in \{1,\dots,c\}^n$. Each data point $i$ is associated with a set of $m$ labeling function outputs $\bm \Lambda_i \in \{0,1,\dots,c\}^m$. A labeling function $j$ votes for a label (e.g $\Lambda_{i}^j=2$) or abstains ($\Lambda_{i}^j = 0$). This means that in this simple multi-class framework we assume that label functions only output positive evidence towards a particular class label (e.g $\Lambda_{i}^j=2$) as opposed to providing information indicating that something is not indicative of a particular class (e.g $\Lambda_{i}^j=-2$).  

As in~\cite{ratner2016data} and \cite{bach2017learning}, we can model the distribution over latent class variables and labeling functions as a factor graph. The density of the joint distribution of $\bm\Lambda$ and $\bm Y$ is:
\begin{equation}\label{eq:condind}
    p_{\bm \theta}(\bm \Lambda,\bm Y) \propto \exp \left( \sum_{j=1}^m\sum_{i=1}^n \theta_j \phi(\Lambda_{i}^j,y_i) \right)
\end{equation}
where the parameter $\theta_j$ models the accuracy of labeling function $\bm \Lambda^{j}$, and for sample $i$ and labeling function $j$ we define the following factor function:
$$\phi(\Lambda^j_{i},y_i)= \begin{cases*}
      1 & if $\Lambda_{i}^j = y_i$\\ 
      0 & if $ \Lambda_{i}^j = 0$ \\
      -1        & otherwise
\end{cases*}.$$
In this model, the assumption is that the outputs of the labeling functions are conditionally independent given the true label. The parameter vector $\bm \theta$ can be learned by  minimizing the negative log marginal likelihood given the observed labeling function outputs  $\bar{\bm\Lambda}$
$$\argmin_{\bm \theta} -\log \sum_{\bm Y}  p_{\bm \theta}(\bm Y,\bar{\bm\Lambda}).$$  
Note that the function is convex  in $\bm \theta$ and can be learned using (stochastic) gradient descent. The gradient for $\theta_j$ is
\begin{equation}\label{gradient}
\sum_{i=1}^n \left( E_{\Lambda,Y \sim \theta}\left[ \phi(\Lambda_{i}^j,y_i)  \right] - E_{Y \sim \theta| \hat{\Lambda} }\left[ \phi(\bar{\Lambda}_{i}^j,y_i)  \right] \right).  
\end{equation}
Sampling is performed during each iteration of gradient descent to approximate the expectations, since the exact computation requires marginalization. The simple model for the multi-class scenario presented in this section is close to the conditionally independent model for binary classification presented in~\cite{bach2017learning}.
\section{Data programming with pairwise feedback}
We assume that users can write \textit{weak pairwise feedback functions} which output an undirected, sparse graph $\bm A$. In the simplest case, users only write functions that output that pairs $i,j$ should be of the same class ($A_{i,j}=1$) and abstain otherwise ($A_{i,j}=0$), resulting in a symmetric matrix $\bm A \in \{0,1\}^{n\times n}$. 

In some cases, users may also be able to gather reliable information about pairs $i,j$ which should not be of the same class ($A_{i,j}=-1$), resulting in $\bm A \in \{-1,0,1\}^{n\times n}$.
Finally, it may be possible to obtain functions outputting values that indicate a strength of belief that some pairs belong to the same or different class. In this paper, we constrain our analysis to the use of noisy pairwise feedback that is available as a discrete signal, i.e. a function outputs that two samples are of the same or different class rather than a probability thereof. We believe that this is the most intuitive and reliable form of pairwise feedback that users can provide.  

Assume we have $p$ separate sources of pairwise feedback indexed by $k$ in the form of sparse, symmetric matrices $\bm A^k \in \{-1,0,1\}^{n\times n}$. We can again model the joint distribution using a factor graph by simply using $\bm A^k$ to define factors amongst pairs of $\bm Y$:
\begin{equation}\label{eq:pairwise}
    p_{\bm \theta,\bm \eta}(\bm \Lambda,\bm Y) \propto \exp ( \sum_{j=1}^m\sum_{i=1}^n \theta_j \phi(\Lambda_{i}^j,y_i) +  \sum_{i=1}^{n}\sum_{j=1}^n \sum_{k=1}^{p}\eta^k A^k_{ij}\mathbbm{1}\left[y_{i}= y_j \right])
\end{equation}
where $\bm \eta$ is a parameter vector modeling the accuracy of the pairwise sources. Notice that we do not model $\bm A^k$ as a variable but merely use it to create factors and learn one accuracy parameter per source $k$. As in \cite{ratner2016data}, we perform inference via gradient descent and use Gibbs sampling to compute expectations of the distributions used in the gradient update. 
\section{Experiments}
To study the general usefulness of pairwise feedback in the programmatic creation of labeled data, we first conduct experiments on synthetic data in order to have full control over the precision and recall of the labeling functions.  We then use the popular 20 Newsgroup dataset to show improvements in performance on real data and to demonstrate the ease with which same-class pairwise feedback can be generated in practice. 
\subsection{Synthetic data}
\subsubsection{Same-class feedback}
\begin{figure}
\centering
\begin{subfigure}[t]{.49\textwidth}
  \centering
  \includegraphics[width=1.0\linewidth]{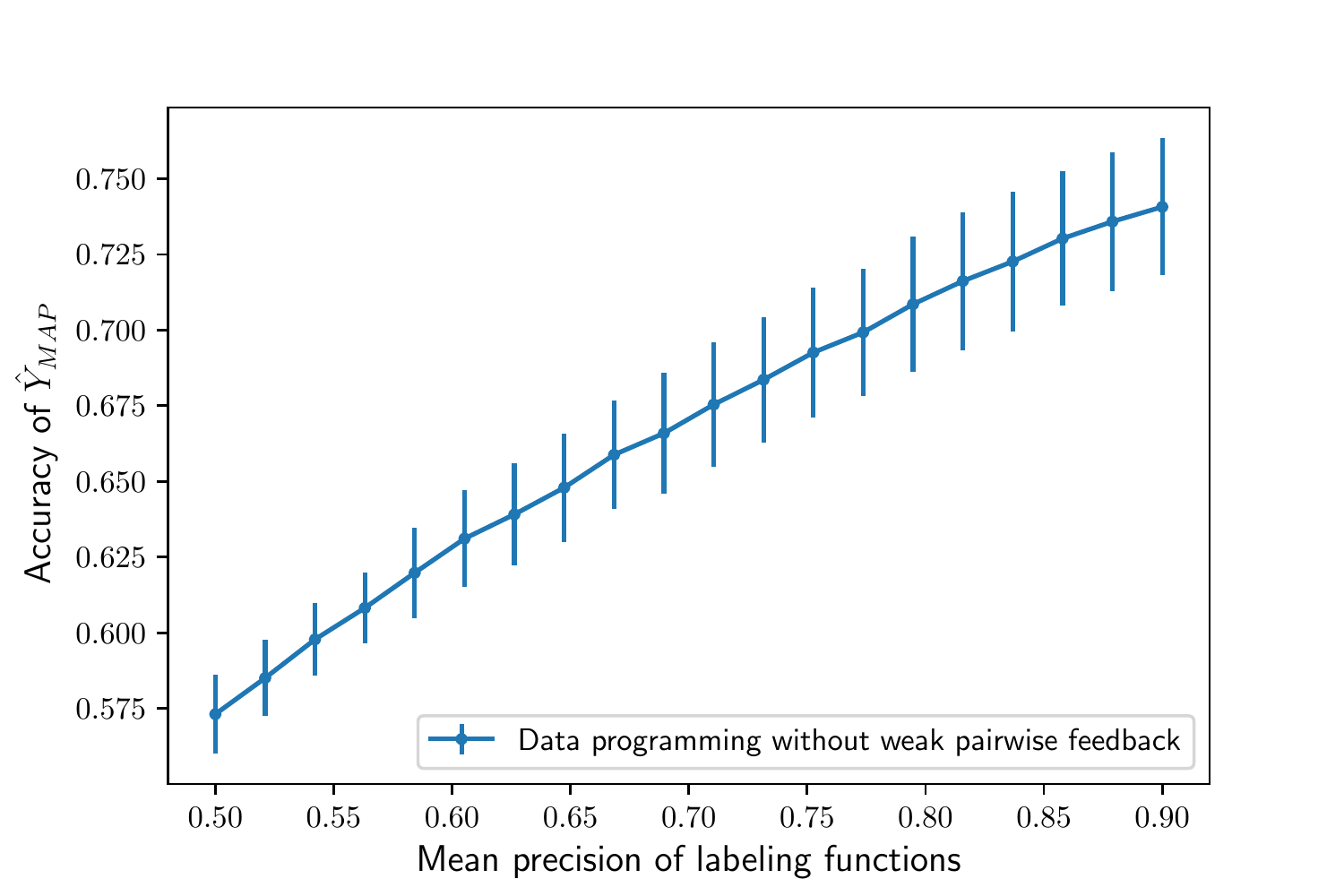}
  \caption{Data programming \textbf{without} pairwise feedback. Y-axis: accuracy of latent class variable MAP estimate. X-axis: average precision of simulated labeling functions at fixed recall.}
  \label{fig:simpleDP}
\end{subfigure}%
\hspace{5pt}
\begin{subfigure}[t]{.49\textwidth}
  \centering
  \includegraphics[width=1.0\linewidth]{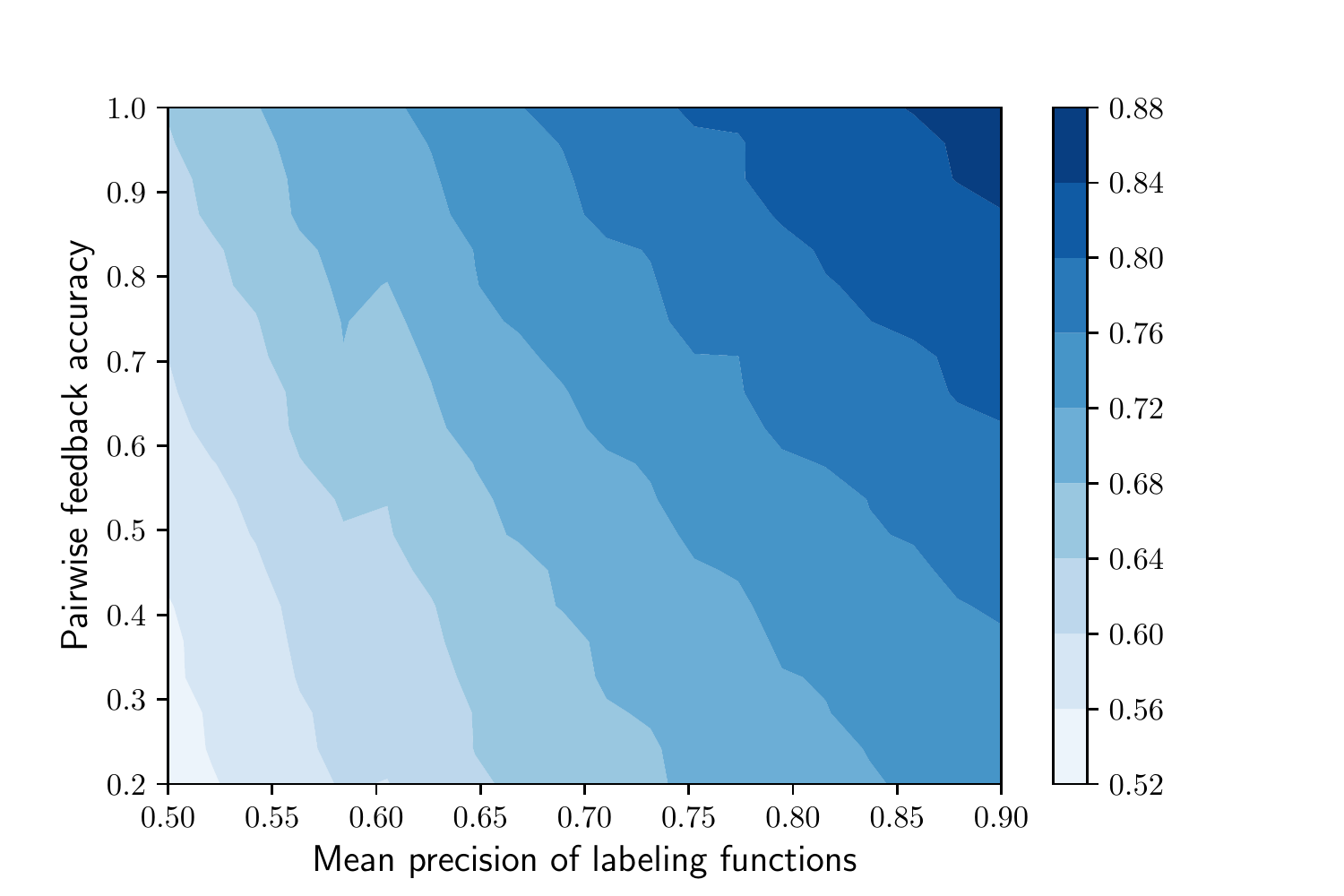}
  \caption{Data programming \textbf{with} pairwise feedback (same-class). Accuracy of $\hat{Y}_{MAP}$ indicated by color bar. Y-axis: accuracy of simulated pairwise feedback ($5000$ pairs). X-axis: average precision of simulated labeling functions at fixed recall.}
  \label{fig:sim_pairwise5kaccuracy}
\end{subfigure}%
\caption{\textbf{Accuracy} of MAP estimate for the latent class variable $Y$ on synthetic data using simulated labeling functions and pairwise feedback.}
\label{fig:accuracy_sim}
\end{figure}
\begin{figure}
\centering
\begin{subfigure}[t]{.49\textwidth}
  \centering
    \includegraphics[width=1.0\linewidth]{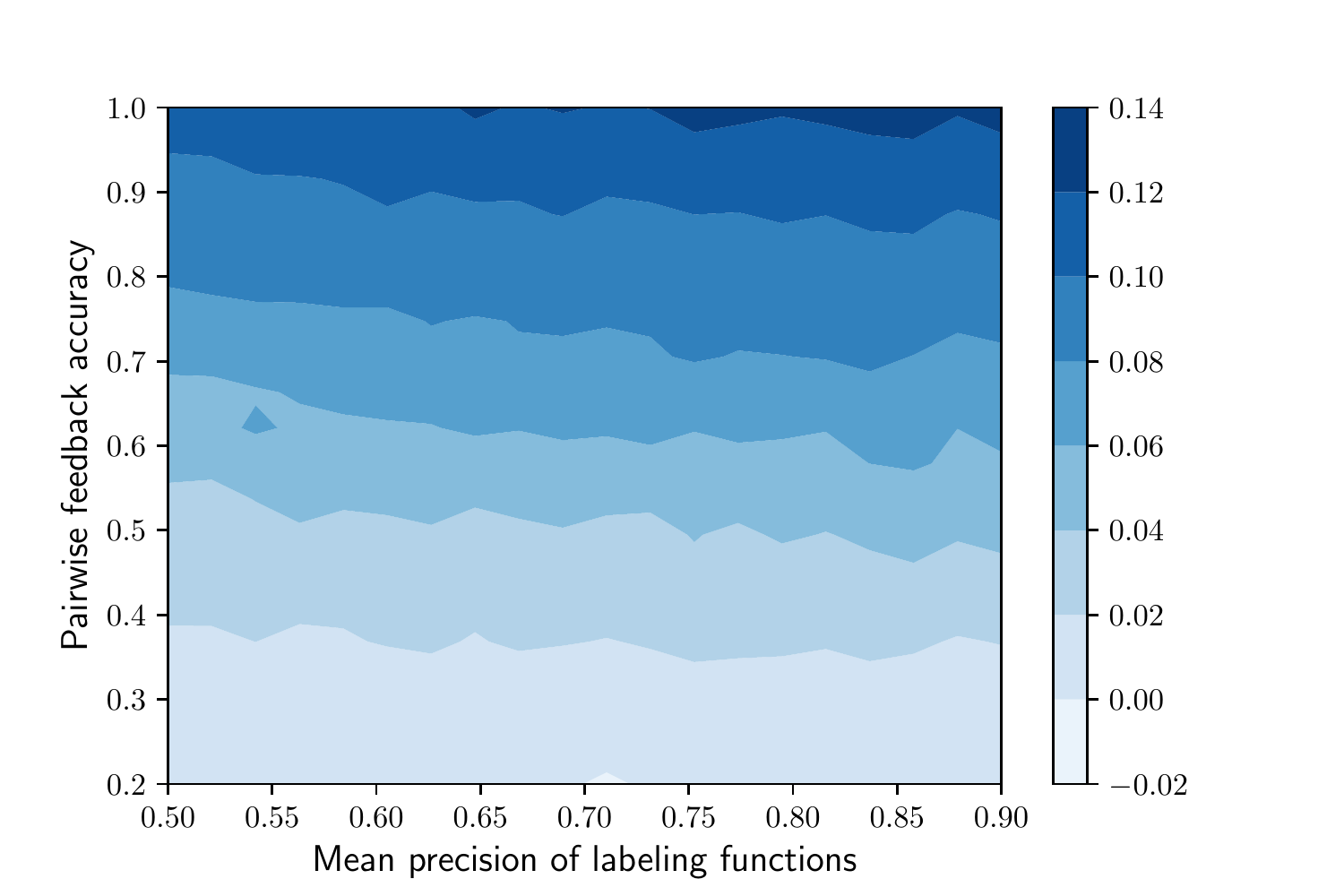}
    \caption{$1000$ pairs.}
    \label{fig:sim_pairwise1kincrease}
\end{subfigure}%
\hspace{5pt}
\begin{subfigure}[t]{.49\textwidth}
  \centering
    \includegraphics[width=1.0\linewidth]{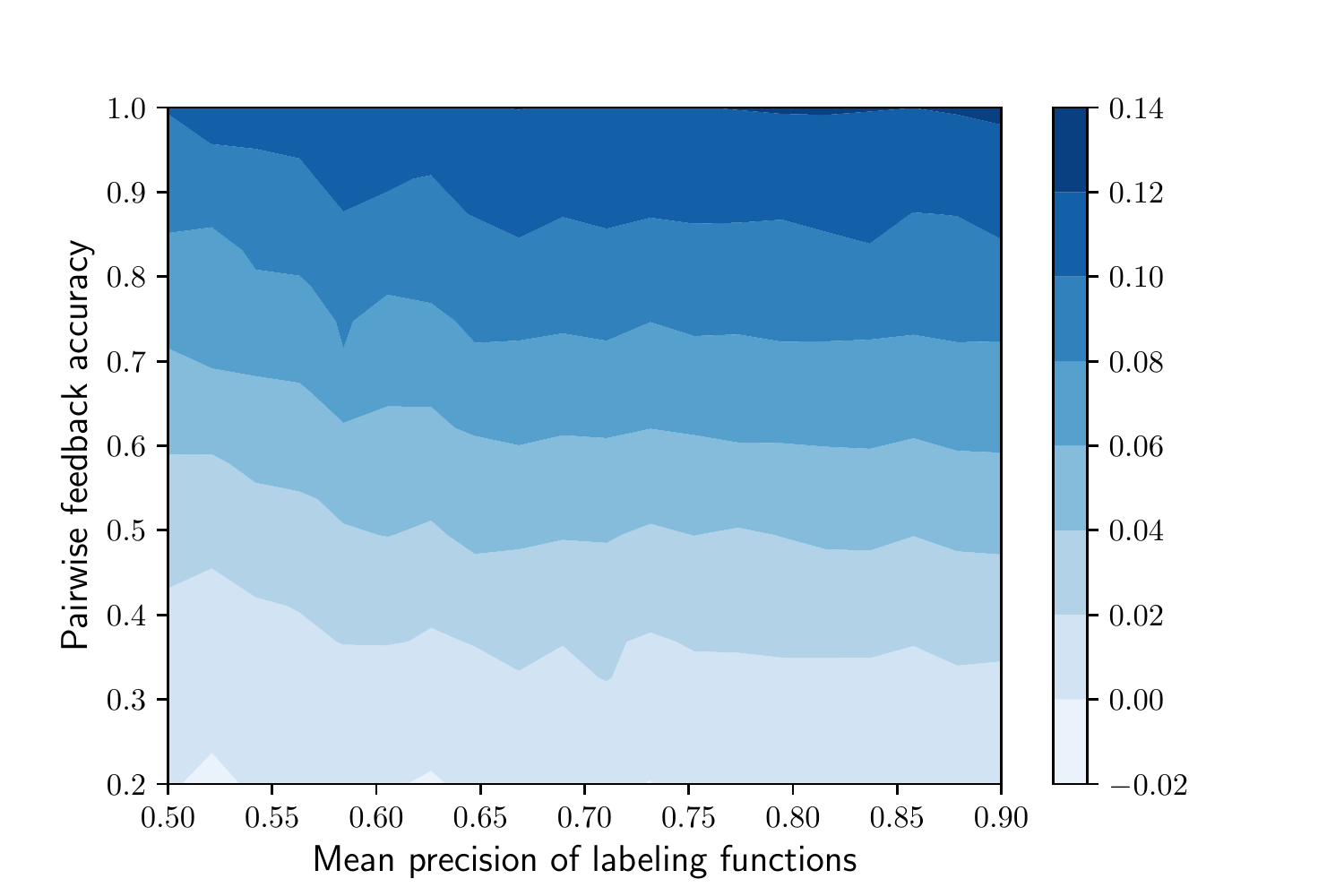}
    \caption{$5000$ pairs.}
    \label{fig:sim_pairwise5kincrease}
\end{subfigure}%
\caption{\textbf{Increase in accuracy} of MAP estimate for the latent class variable $Y$ achieved by data programming \textbf{with} pairwise feedback, compared to a model without pairwise feedback, on synthetic data. Results shown for $1k$ and $5k$ pairs. The increase in accuracy is indicated by the contours.}
\label{fig:accuracy_incrase}
\end{figure}
We simulate a multi-class classification task with $c=10$ classes and generate two weak labeling functions per class. For each weak labeling function, we randomly add false positives (fp) and false negatives (fn) to the true label to achieve a target recall and precision.  
Given a target mean recall and precision, the specific recall and precision for one run of an experiment are drawn randomly from a truncated normal distribution. Throughout the experiments, we fix the target recall of each labeling function at $0.5$ and increment the target precision from $0.5$ to $0.9$. 
To create pairwise feedback, we generate same-class information $A_{i,j}=1$ by first specifying a target count of pairs and a target accuracy and then randomly create fp and true positive (tp) pairs. To provide a point of reference for the quality of the simulated pairwise feedback one can consider that for a dataset with $c$ classes and even class balance, the default accuracy of a pairwise matrix indicating same-class membership will be $\sim 1/c$. 

We repeat this process of simulating labeling functions and weak pairwise feedback ten times and present results based on the mean accuracy achieved by the MAP estimate $\hat{Y}_{MAP}$. Figure~\ref{fig:simpleDP} shows the accuracy of $\hat{Y}_{MAP}$ produced by a data programming model with independent labeling functions (Equation~\ref{eq:condind}) on the simulated task as the precision of the underlying labeling functions increases. The contour plot in Figure~\ref{fig:sim_pairwise5kaccuracy} shows the accuracy of $\hat{Y}_{MAP}$ when $5000$ pairs of \textit{same-class feedback} of varying accuracy are added to the model (Equation~\ref{eq:pairwise}). Figure~\ref{fig:sim_pairwise5kincrease} shows the increase in $\hat{Y}_{MAP}$ accuracy gained by using this weak pairwise feedback. Figure~\ref{fig:sim_pairwise1kincrease} reveals that even a small number of $1000$ pairs can lead to drastic improvements.
\subsubsection{Same-class and different-class feedback}
\begin{figure}[h]
    \centering
    \includegraphics[width=0.7\linewidth]{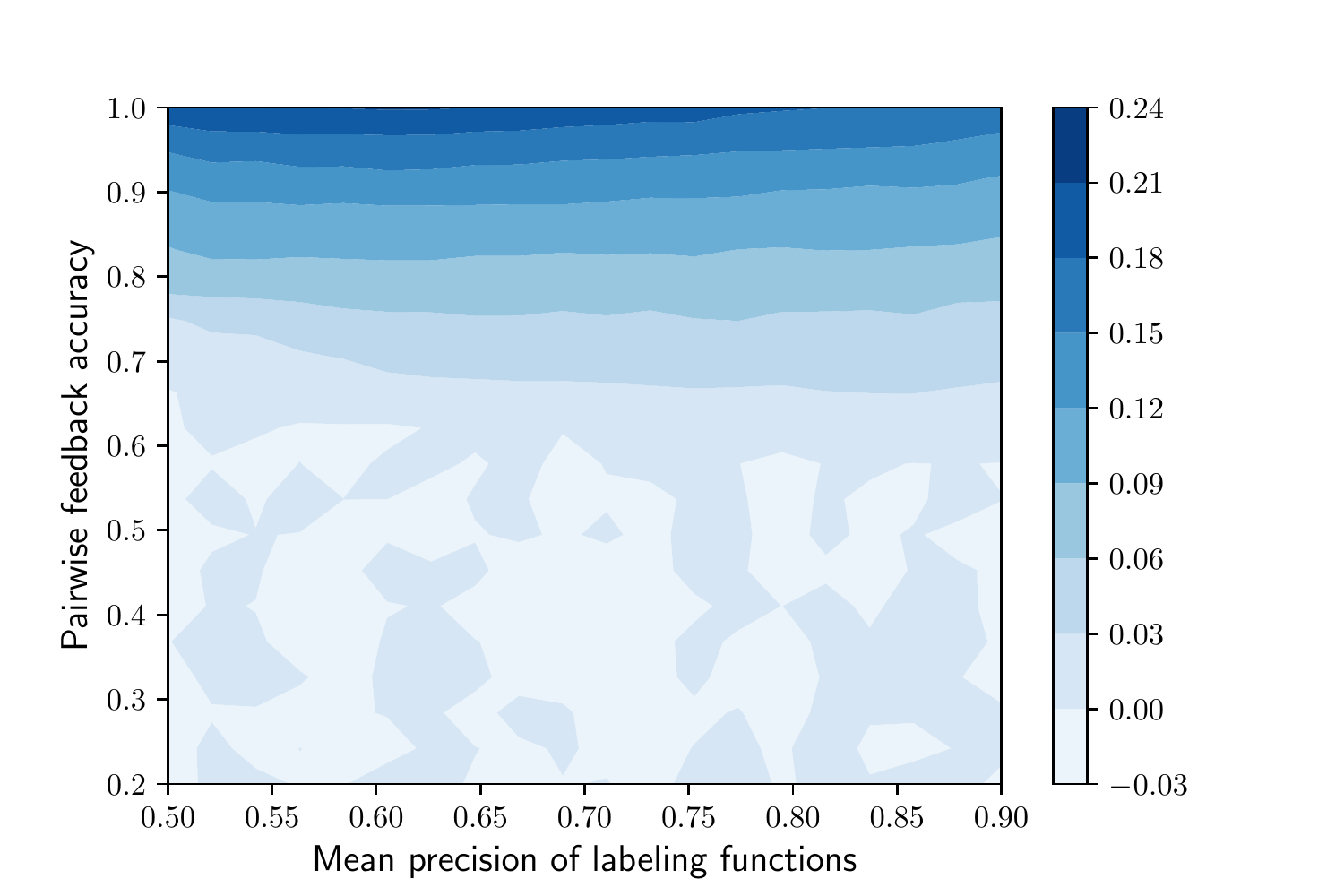}
    \caption{\textbf{Increase in accuracy} of MAP estimate for the latent class variable $Y$ achieved by data programming \textbf{with same-class and different-class pairwise feedback}, compared to a model without pairwise feedback, on synthetic data. Results shown for $10k$ pairs. The increase in accuracy is indicated by the contours.}
    \label{fig:pairwise_negative}
\end{figure}
Using the same approach as described in the previous section, we also simulated the acquisition of different-class feedback $A_{i,j}=-1$ in addition to same-class feedback. To this end, we randomly sampled unique pairs from the data and then corrupted the ground-truth pairs to achieve desired noise levels of pairwise feedback. Results for an experiment with $10,000$ pairs are shown in Figure~\ref{fig:pairwise_negative}. Comparing  Figure~\ref{fig:pairwise_negative} to Figure~\ref{fig:accuracy_incrase}, it appears that a higher pairwise accuracy is needed to achieve consistent improvements when pairwise feedback contains both same and different class feedback, as the negative feedback in this setting is naturally less informative. 
\subsection{Newsgroup data}
\begin{table}[ht]
    \centering
    \begin{tabular}{l|rr}
             Source &  Unique pairs  &  Pair accuracy \\
            \midrule
            Email pairs &         21696 &                          87.33\% \\
            MKNN pairs &         12355 &                         88.04\% \\
            \bottomrule
            \end{tabular}
    \caption{Weak pairwise sources created for a subset of 20 Newsgroup topics. }
    \label{tab:pairmatrices}
\end{table}
\begin{table}[ht]
    \centering
    \begin{tabular}{r|rrl}
 Label function &  Precision &    Recall &               Class \\
\midrule
              1 &   0.896970 &  0.185232 &         alt.atheism \\
              2 &   0.916216 &  0.424280 &         alt.atheism \\
              3 &   0.934263 &  0.474696 &      comp.windows.x \\
              4 &   0.646739 &  0.120445 &      comp.windows.x \\
              5 &   0.738806 &  0.100202 &      comp.windows.x \\
              6 &   0.650000 &  0.039514 &           sci.space \\
              7 &   0.835635 &  0.612969 &           sci.space \\
              8 &   0.821739 &  0.191489 &           sci.space \\
              9 &   0.818966 &  0.096251 &           sci.space \\
             10 &   0.725118 &  0.310030 &           sci.space \\
             11 &   0.898734 &  0.071935 &           sci.space \\
             12 &   0.940741 &  0.257345 &           sci.space \\
             13 &   0.617801 &  0.152258 &  talk.politics.misc \\
             14 &   0.765152 &  0.130323 &  talk.politics.misc \\
             15 &   0.787709 &  0.181935 &  talk.politics.misc \\
             16 &   0.804878 &  0.052548 &  talk.religion.misc \\
    \bottomrule
    \end{tabular}
    \caption{Weak labeling functions created for a subset of 20 Newsgroup topics.}
    \label{tab:labelfunc}
\end{table}
We used a subset\footnote{topics: alt.atheism, talk.religion.misc, talk.politics.misc, comp.windows.x, sci.space} of the popular 20 Newsgroups text classification dataset\footnote{As available at:\\ \url{https://scikit-learn.org/stable/datasets/index.html\#newsgroups-dataset}} with  $c=5$ classes resulting in $n=4177$ samples and roughly even class balance. 
For this subset, the default accuracy of a classifier predicting the majority label is $23.65\%$. We manually created 16 labeling functions based on user-defined heuristics that look for mentions of specific words, see Table~\ref{tab:labelfunc} for details regarding precision and recall.

To demonstrate the ease with which pairwise same-class feedback can be obtained we create two different sources thereof, one using meta-data and one based on MKNN. We use a regular expression to extract the first Email mentioned in each document. Using extracted Emails, we create pairs of documents $i,j$ containing the same Email address and assign a label $A^1_{i,j}=1$. The assumption is that the first email in a document identifies the author, and that documents by a unique author are likely to belong to the same topic. This results in a pairwise matrix $\bm A^1$  covering about $0.12\%$ of all possible pairs, see Table \ref{tab:pairmatrices}. 

Next, we create a pairwise matrix based on MKNN, a popular heuristic that works well in many domains (see e.g. \cite{fogel2019clustering} for its use on image data). We use a term frequency - inverse document frequency (tf-idf)\footnote{With a minimum document frequency cutoff at $0.001$.} embedding and compute the 10 nearest neighbors for all documents using cosine similarity. Using this nearest neighbor graph, we subsequently create pairs of documents $i,j$ that are MKNN\footnote{$i$ is amongst the 10 nearest neighbors of $j$, and $j$ is amongst the 10 nearest neighbors of $i$.} and assign a label $A^2_{i,j}=1$. Note that both heuristic sources we use only cover same-class pairs. 
\begin{table}
    \centering
    \begin{tabular}{l|rr}
    & \shortstack[l]{Mean accuracy of\\ posterior class label estimate} &  Standard deviation \\
    \midrule
    Majority vote & 46.94\% &  0.000 \\
    Labeling functions &  56.45\% &            0.005 \\
    + Email pairs &  59.64\% &            0.006 \\
    + MKNN pairs &  75.98\% &            0.008 \\
    + MKNN + Email pairs &    69.92\% &            0.009 \\
    \bottomrule
    \end{tabular}
    \caption{Results obtained for modeling of the latent class variable for a subset of the 20 Newsgroup dataset.}
    \label{tab:newsgroupresults}
\end{table}

As a comparison baseline, we use the conditionally independent model in Equation~\ref{eq:condind} as well as a majority vote using the weak labeling functions. Results are shown in Table~\ref{tab:newsgroupresults}, demonstrating the increase in accuracy of modeling the latent class variable when pairwise feedback is incorporated.

\section{Discussion}
\begin{figure}
    \centering
    \includegraphics[width=0.7\textwidth]{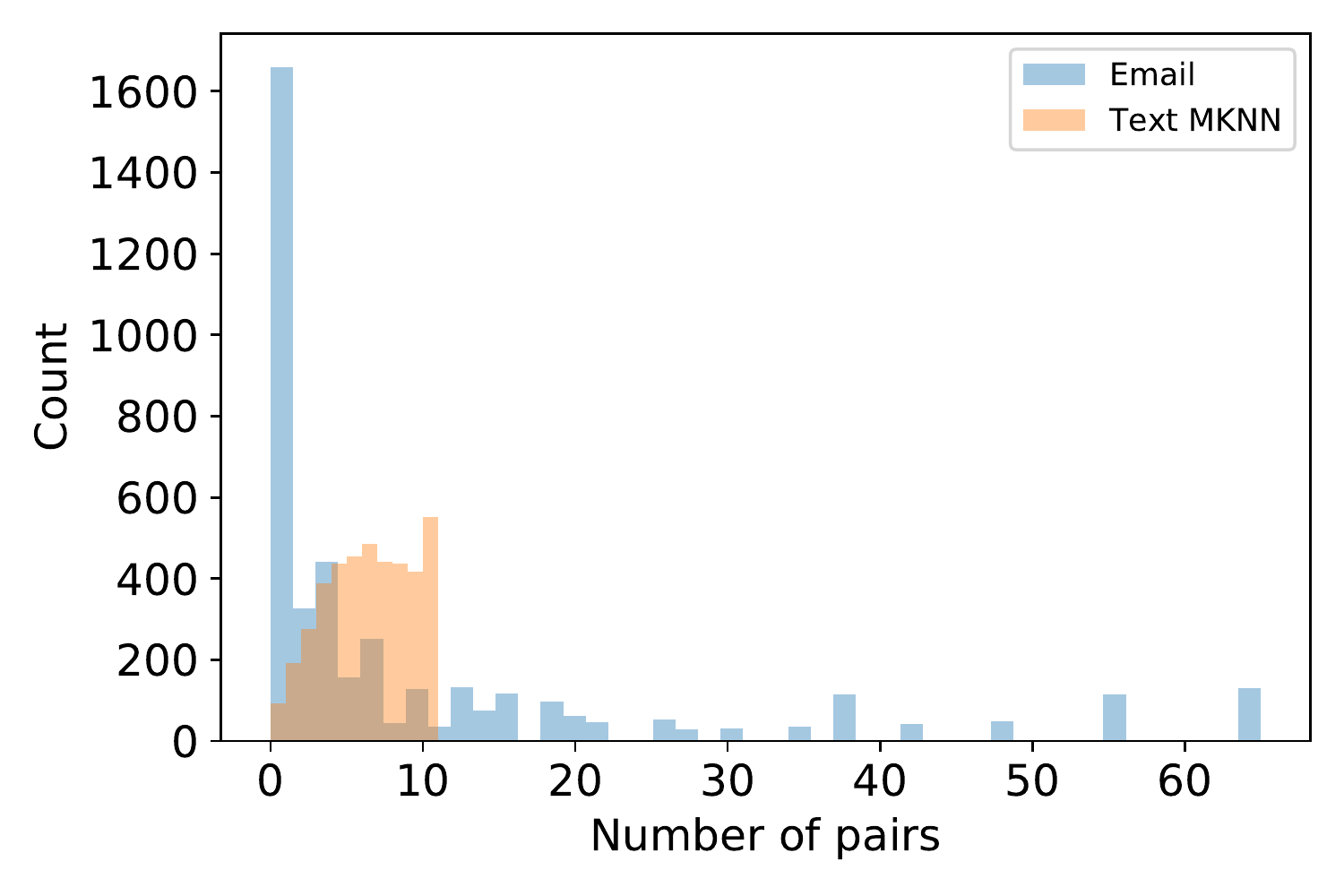}
    \caption{A histogram showing the distribution of pairs per sample in the 20 Newsgroup dataset for the two different pairwise matrices that were created. The plot shows that the text-based MKNN matrix pairs are by construction less biased across the dataset compared to the pairwise matrix created from matching emails.}
    \label{fig:pairhistogram}
\end{figure}
Our experiments on synthetic and benchmark data show that even a single, small source of weak pairwise feedback can lead to substantial improvements in the posterior estimate of the latent class variable.  Our experiment on the 20 Newsgroup data demonstrates the ease with which good sources of pairwise feedback can be obtained for common classification tasks. The difference in performance of the Email pairs compared to the text-based MKNN pairs is very likely due to the less biased pairwise information the MKNN matrix presents, see Figure~\ref{fig:pairhistogram}. The experiment demonstrates that improvement in performance is not just a function of the accuracy of pairs obtained from a feedback source, but also of the distribution of this information across the samples. In the experiment, the MKNN pairs tie more evidence of the labeling functions together across the entire dataset compared to the Email pairs.

We emphasize that the performance of the model without pairwise feedback is not a shortcoming of data programming by any means. Without weak labeling functions, pairwise feedback by itself would not allow us to model the latent class structure. Furthermore, only a small amount of human labeling effort was required in creating the weak labeling functions and no outside source of information was used. More time as well as knowledge about the domain can most certainly lead to additional, high quality labeling functions that could improve performance of the model that does not use pairwise feedback. Our experiments reveal the usefulness of the programmatic creation of labeled datasets and that the performance can be increased by additionally gathering weak pairwise feedback.

\section{Conclusion}
We have empirically demonstrated that just one noisy pairwise function can substantially improve accuracy of the posterior estimate of the latent class variable in data programming. Noisy pairwise feedback is a valuable resource for data programming as it ties evidence of labeling functions together, across different samples of a dataset.  Our experiments as well as related literature (e.g. in semi-supervised clustering) demonstrate that pairwise feedback with sufficient accuracy can be collected with ease and at scale for many problem types. 

Several issues remain that we aim to address in our future work. We will expand the number of experiments on real world data to comprehensively validate current claims, and include weak sources indicating different-class membership.  Furthermore, we will explore alternative inference algorithms. As the number of samples grows, even a small fraction of all possible pairs can become quite large. We also intend to conduct a theoretical analysis of the expected improvements and guarantees when pairwise feedback is used in the programmatic creation of labeled data.




\subsubsection*{Acknowledgments}
This work has been partially supported by DARPA FA8750-17-2-013 and by a Space Technology Research Institutes grant from NASA’s Space Technology Research Grants Program.


 \bibliographystyle{apalike}
\bibliography{mybib}
\end{document}